# Visual Sensation and Perception Computational Models for Deep Learning: State of the art, Challenges and Prospects

Bing Wei, *Member, IEEE*, Yudi Zhao, Kuangrong Hao[*], and Lei Gao, *Member, IEEE*

*Abstract*—**Visual sensation and perception refers to the process of sensing, organizing, identifying, and interpreting visual information in environmental awareness and understanding. Computational models inspired by visual perception have the characteristics of complexity and diversity, as they come from many subjects such as cognition science, information science, and artificial intelligence. In this paper, visual perception computational models oriented deep learning are investigated from the biological visual mechanism and computational vision theory systematically. Then, some points of view about the prospects of the visual perception computational models are presented. Finally, this paper also summarizes the current challenges of visual perception and predicts its future development trends. Through this survey, it will provide a comprehensive reference for research in this direction.**

## I. INTRODUCTION

EMBRACED by enormous amounts of incoming information from the visual environments, the human visual system (HVS) can cope with the situation effortlessly with limited computational resources and capacity [1], [2]. During visual information processing, visual sensation and perception is the ability to interpret the surrounding environment using light in the visible spectrum reflected by the object in the environment [3]. Benefiting from the neurophysiologists' research for the visual system, we have a preliminary understanding of how the biological visual perception system performs visual activities, such as information acquisition, vision processing, and vision understanding [4]. However, the structures and functions of visual perception are extremely complex and diverse. Recently, the research shows that the visual structures involved in the process of visual sensation and perception mainly include the retina, the lateral geniculate nucleus, and visual cortex [5]. Meanwhile, the biological perception characteristics of the whole visual system mainly include visual attention, visual memory, inference perception, feedback, spatial-time coordinated perception, specificity, sparse response, and transfer, as shown in Fig. 1. So far, the high-level research on the structure and mechanisms of the biological visual perception system has not made breakthrough progress, and the fundamental theoretical research on visual perception system requires some new technologies and methods. However, the characteristics of biological visual perception, such as robustness, computational efficiency, and rapid response, have important inspirations for the design of artificial intelligence algorithms and structures.

In recent years, artificial intelligence (AI) can significantly benefit from the ongoing research in neuroscience and psychology. Looking into these fields as a source of inspiration will pave the way for building learning machines that mimic biological structure and mechanisms. Turing machine and perceptron are the earliest perceptual computing models, which are used to mimic the perceptual computing ability of the biological brain. For instance, convolutional neural network (CNN) [6] inspired by studying a cat's visual cortex can be viewed as a success story in following this line of thinking. At present, the important deep learning models include recurrent neural network (RNN) [7], auto-encoder (AE) [8], [9], generative adversarial network (GAN) [10], graph neural network (GNN), and reinforcement learning models. These network models can achieve good performance on some tasks, but do not consider the mechanisms and structures of biological visual perception and complete higher-level intelligent computing.

In addition, combining these deep learning models with biological visual perception, many biologically plausible approaches to deep learning have been proposed for reviews. Existing approaches usually use either involved architectures or elaborate mechanisms to approximate the backpropagation algorithm. Many intelligent computing models based on biological visual perception have been proposed to complete the practical tasks, such as object detection [11], image recognition [12], natural language processing [13], semantic segmentation [14], series data prediction [15], [16], gesture recognition [17], and so on. Meanwhile, visual perception computational models oriented deep learning have developed into an interdisciplinary discipline involving many deep learning networks such as supervised learning, semi-supervised learning, self-supervised learning, unsupervised learning.

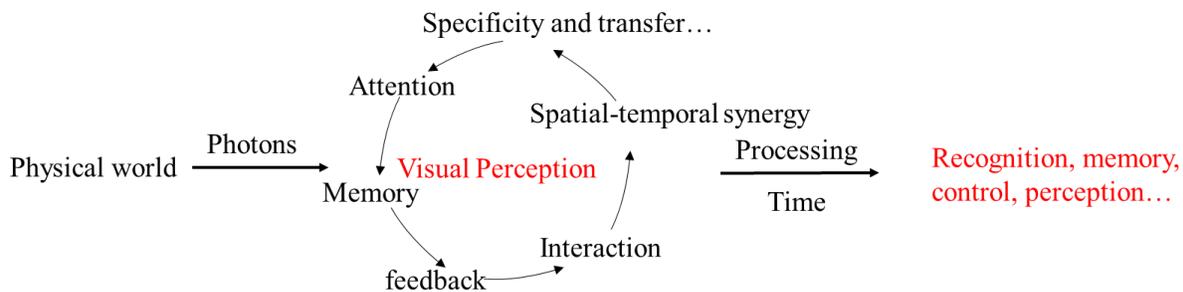

Fig. 1 Biological perception processes with different perception characteristics

Inspired by the structure, characteristics and mechanisms of the biological visual perception system, its many difficulties and problems in artificial intelligence and deep learning models can be solved, such as fastness, robustness, feature distinguishability, and neural network structure. Meanwhile, how to intimate these mechanisms can further provide a kind of bridge between deep learning and biological visual system is a very attractive topic. In this survey, we take the fundamental idea coming from visual sensation and perception and their application in deep learning. We first introduce the recent progress of visual perception computational models oriented deep learning, showing its advantage and its progressive impact on artificial intelligence. The outline of the contributions of this paper can be summarized as follows:

(1) Comparisons with other survey papers in the fields of visual perceptions have been accomplished, this survey summarizes the latest visual perception computational models oriented deep learning for the first time to our knowledge, including not only research on a single deep learning model including CNN, RNN, AE, GAN, GNN, but also reinforcement learning models shown in Fig. 2.

(2) Some limitations and the main challenges faced by current visual perception methods have been introduced from the perspective of biological visual perception and artificial intelligence.

(3) Some development prospects of visual sensation and perception have been put forward to reflect the theme of the paper, for which researchers in this field have pointed out the direction of scientific research.

The rest of this paper is organized as follows. Section II introduces some visual perception computational models for deep learning. Section III elaborates some challenges faced by visual perception computational models for deep learning we can see now. Section IV illustrates its development prospects and trends, and Section V concludes this paper.

## II. VISUAL PERCEPTION COMPUTATIONAL MODEL FOR DEEP LEARNING

The benefits to develop visual perception computational models oriented artificial intelligence are two-fold. First, biological visual perception provides a rich source of inspiration for new types of algorithms and architectures. Furthermore, it can be independent and complement to the mathematical and logic-based methods that have largely dominated traditional approaches to deep neural networks. Second, visual perception can provide validation of deep neural network techniques that already exist. The main application areas of visual perception computational models oriented artificial intelligence are shown in Fig. 2. In this section, we introduce some visual perception computational models, including visual attention based, memory-based, inference perception-based, feedback-based, multi-modal-based, spatial-time synergy-based, cross-level and interaction based, specificity and transfer based, and sparse response based. These algorithms are hot spots for researchers and basically cover different intelligent visual perception processes.

Identifying some of these problems will help in reunifying natural and artificial vision and addressing more challenging questions as needed for building adaptive and versatile integrated artificial systems which are deeply bio-inspired.

### A. Spatio-Temporal Cooperation

#### 1) Attention-based Computational Models

Visual attention refers to the cognitive operations of selectively focusing on crucial information and simultaneously filtering out irrelevant information from visual scenes [18], [19]. In the visual attention process, bottom-up mechanism or top-down mechanism can be caught by external environment stimuli or driven by human internal goals, respectively [20]–[22]. Inspired by visual attention mechanisms, computational models can effectively select information and improve the accuracy of data processing [23]. Visual attention based recurrent network has achieved great performance in image classification [24] and multiple object recognition [25]. Not only that, attention mechanism has been widely applied to various deep learning models, which have achieved great success in computer vision and natural language processing (NLP) as well as other tasks.

Based on biological mechanisms, visual attention mechanisms are mainly divided into bottom-up attention and top-down attention [26], and based on theses two kinds of attention, many researches have been done. Chen et al. [11] integarated a top-down multi-modal fusion network with an attention-aware cross-modal cross-level fusion (ACCF) block, which successfully solve the problems of RGB-D salient objection detection. Anderson et al. combined both the bottom-up and top-down attention with Faster R-CNN, where bottom-up mechanism gives image regions with associated feature vectors and top-down mechanism determines feature weights [27]. Besides, Haut et al. [26] introduced a visual attention based ResNet for hyperspectral image (HSI) classification by simulating both the bottom-up and the top-down attention selection of visual cortex.

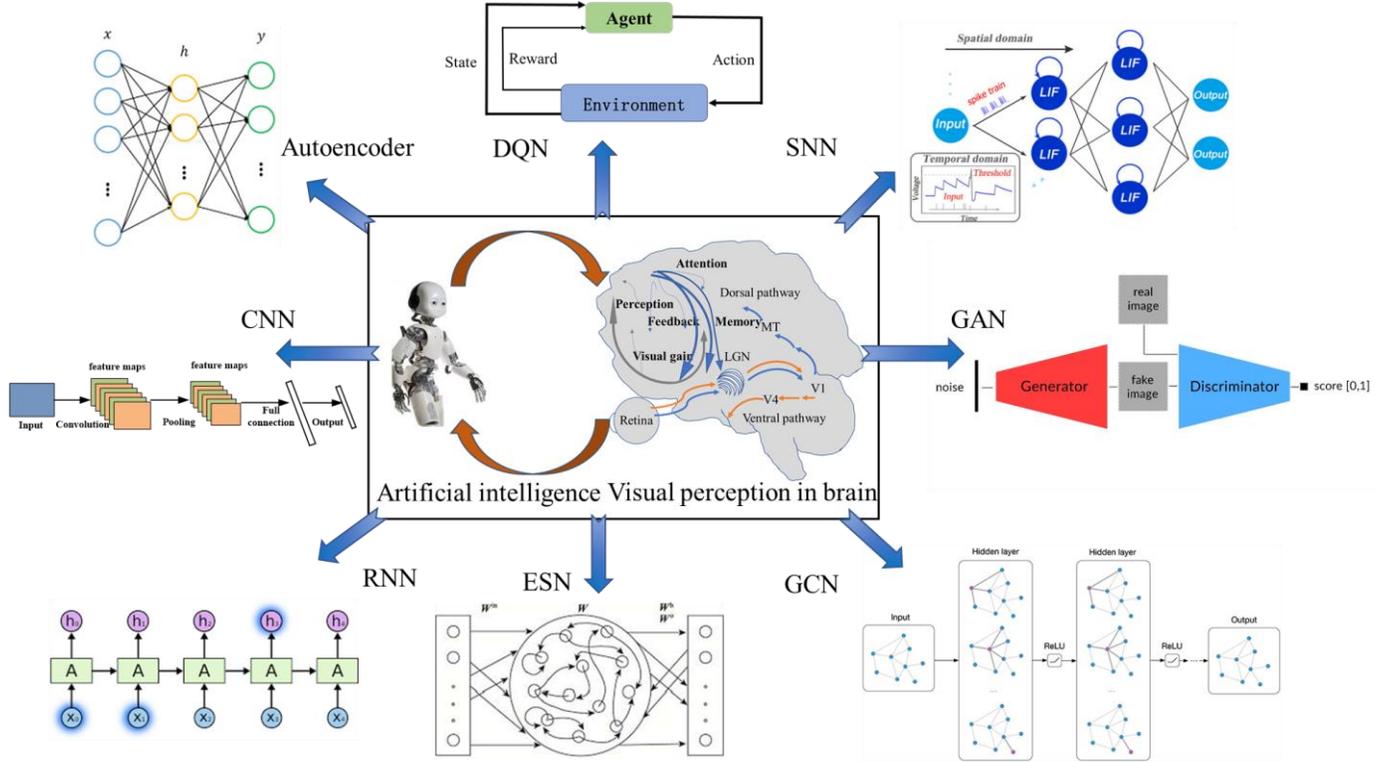

Fig. 2. Visual perception with deep learning models

Apart from the visual attention mentioned above, there also exists researches that foucus on channel-wise attention and spatial attention. Abdelhalim et al. [28] employed channel-wise self-attention in generative adversarial networks (GANs) for skin lesion data augmentation. Attetion GANs based method [29] was also proposed for aerial scene classification, where the attention module was divided into the channel attention and the spatial attention. Besides, Chen et al. [30] incorporate channel-wise attention into a CNN in addition to spatial attention, which helps notice each feature entry in the multi-layer 3D feature maps. Furthermore, channel and spatial attention is also utilized to enhance the common semantic information [31] and detect crack [32]. To alleviate the negative affects of mislabeled data, Min et al. [33] design a new Mutually Attentive Co-training Framework (MACF), which considered the spatial and channel attetnion.

Especially, there is also a consititude of the top-down attention and the channel-wise attention in a three-stream attention-aware cross-modal cross-level combination (Att-CMCL) block for RGB-D salient objection detection [34]. Here, we summarize the representative computational models based on visual attention mechanism in Table 1.

Table 1. Computational models based on visual attention mechanism

| Methods | Mechanisms | Task description | Learning type |
|---|---|---|---|
| RAM [24] | RNN + Attention | Image Classification | supervised |
| DRAM [25] | DRNN + Attention | Multiple Object Recognition | supervised |
| ACCF Block [11] | ConvNets + Top-down Attention | RGB-D Salient Objection Detection | supervised |
| RCNN-BATA [30] | R-CNN + Bottom-up Attention + Top-down Attention | Image Captioning + Visual Question | supervised |
| A-ResNet [26] | ResNet + Bottom-up Attention + Top-down Attention | HSI Classification | supervised |
| SPGGANs [28] | GANs + Channel-wise Self-attention | Data Augmentation for skin Lesion | unsupervised |
| Attention GANs [29] | GANs + Channel Attetnion + Spatial Attention | Aerial Scene Classification | unsupervised |
| SCA-CNN [30] | CNN + Channel Attetnion + Spatial Attention | Image Captioning | supervised |
| CSA [31] | DNN + Channel Attetnion + Spatial Attention | Deep Object Co-segmentation | supervised |
| VGG-CASA [32] | VGG + Channel Attetnion + Spatial Attention | Crack Detection | supervised |
| MACF [33] | CNN + Channel Attetnion + Spatial Attention | Classification + Biomedical Segmentation | supervised |
| Att-CMCL Block [34] | VGGNet + Top-down Attention + Channel-wise Attention | RGB-D Salient Objection detection | supervised |

### 2) Visual Memory based Computational Models

Visual memory plays a critical role in senior cognitive functions, since visual memory is the received visual information which is encoded and then restored in the brain, and finally can be used to recall, recognize, categorize and identify the objects [35], [36]. Inspired by the whole visual memory process, Dai et al. [37] simulated the visual memory as a feature learning and feature imagination (FLFI) process, and hence constructioned a bottom-up feature learning and a top-down feature imagination for object detection. Besides, to imitate the procedure when human observers and recognize images, Akagunduz et al. [38] defined this process as the accumulated memorable parts of an image and put forward a novel concept named Visual memory schema (VMS).

Apart from the aspect of visual memory's formation process , visual memory is usually subdivided into three subsystems: visual sensory memory, visual short-term memory (VSTM), and visual long-term memory (VLTM). Since visual sensory memory plays little or even no role in human cognition [36], VSTM and VLTM are mostly researched in visual memory based computational models. VSTM contains a small amount of information which is sparse and coarse due to the limit of storage capacity. However, VLTM has a relatively larger capacity with abundant, detailed, and specific features [39]. Currently, VSTM and VLTM related researches have become a hotspot in artificial intelligence field. Ghosh el al. [40] aimed at developing a short-term memory (STM) based deep brain learning network (DBLN) with two error feedback loops for shape-reconstruction. A multimode long short-term memory network (LSTM) in the three-dimensional space is designed for dense pedestrian tracking, perfectly addressing the uncertainty in trajectory segments length and interval [41]. Motivated by human visual perception and visual memory mechanism, Zhao et al. [42] proposes a visual long-short-term memory (VLSTM) based integrated CNN model for extracting visual perception (VP) information, VSTM information, and VLTM information of fabric defect images.

In addition to these mentioned above, Song et al. [43] proposed a local-global memory neural network (LGMNN) model ,where the local memory helps learn the individual patterns of a patient and the global memory can learn the group information of disease under interindividual relationshipsm, this model can learn the hidden knowledge of electronic medical records.

*3) Spatio-Temporal Coordination based Computational Models*

Visual attention and visual memory have been applied to deep learning models successfully but each is in view of their own mechanisms. However, the human visual perception system is a complex system and exists various kinds of cooperation among subsystems. In this section, we consider the ideas inspired by collaboration of spatial information and temporal information in deep learning models [44]. Although they belong to different dimensions but are processed and encoded collaboratively for human perception and recognition [45], [46]. Here, spatial information and temporal information denote a broad sense. For example, visual attention can be considered as a kind of spatial information, whereas visual memory can be categorized to temporal information.

Recently, spatio-temporal coordination mechanism has been applied to video data successfully. For instance, sun et al. [47] combined the motion information on the space aix with the memory information on the time axis of in a step-gained fully convolutional network (SGF) for video saliency detection, which is inspired by the memory mechanism and the visual attetnion mechanism of human being at times of watching a video. [48], [49] also explored and encoded spatial and temporal features of videos, where [48] was for Video Object Segmentation and [49] was for Video Object Segmentation. However, Yan et al. [50] took full advantages of just visual attention mechanism of human beings and proposed a novel spatial-temporal attention mechanism (STAT) for video captioning, where both the spatial information and temporal information come from attention information.

Apart from the video related applications, researchers also utlize the spatio-temporal coordination mechanism in the traffic field. Chen et al. [51] propose a brain-inspired cognitive model mimicking human visual cortex with attention mechanism and long-short memory, which is successfully applied to basic self-driving tasks with only cameras. To extract the spatial and short-term temporal information of traffic flow data, a hybrid deep learning model [52] with attention-based Con-LSTM networks is given, as well as a bidirectional LSTM (Bi-LSTM) for further exploring long-term temporal features.

Furthermore, Yuan et al. [53] considered the sptiotemporal quality-relevant interactions in soft sensor modeling, and proposed a spatiotemporal attention-based LSTM (STA-LSMT) model. For sentiment analysis, Ehsan et al. [54] designed an attention-based bidirectional CNN-RNN deep model (ABCDM) for extrating both the termporal information and the attention information by using bidirectional LSTM and gate recurrent unit (GRU). To capture the spatio-temporal relationships at different times, Liu et al. [55] designed a dual-stage two-phase (DSTP) model for long-term and multivariate time series prediction, which can refer to various fields, such as finance, medicine, energy and environment.

It should be noted that, most of the existing methods based on collaborative spatio-temporal information can't represent the mechanism we raise, which is inspired by spatio-temporal coordination mechanism of human beings. In this section, the idea we want to express includes but not limits to the researches mentioned above. Here we just show enlightenment and hope to discover and digging more between spatio-temporal coordination mechanism and deep learning.

Table 2. Spatio-temporal coordination based computational models.

| Methods | Mechanisms | Task description | Learning type |
|---|---|---|---|
| SGF [47] | VGGNet + Spatial-temporal Information | Video Saliency Detection | supervised |
| ConvGRU [48] | ConvGRU + Spatial-temporal Information | Video Object Segmentation | supervised |
| HPNet [49] | DCNN + Spatial-temporal Information | Video Prediction | supervised |
| STAT [50] | ConvNet + Spatial-temporal Attention | Video Captioning | supervised |
| CMA [51] | CNN + Spatial-temporal Information | Self-driving | supervised |
| Con-LSTM [52] | Con-LSTM + Spatial and Short-term temporal Information | Traffic Flow Prediction | supervised |
| STA-LSTM [53] | LSTM + Spatial-temporal Attention | Industrial Soft Sensor Modeling | supervised |
| ABCDM [54] | CNN-RNN + Spatial-temporal Information | Sentiment Analysis | supervised |
| DSTP-RNN [55] | RNN + Spatial-temporal Information | Time Series Prediction | supervised |

### B. Cross-Level Feedback and Interaction

#### 1) Cross-Level Feedback

Sufficient evidence has proved the complex feedback networks in the brain by neuroscience researches [56]–[58], which means that visual feedback mechanism plays a significant role in processing visual information flexibly and robustly[59], [60]. In this section, we summarize the utilization of visual feedback mechanism in deep learning models from different aspects.

The top-down feedback mechanism has been employed in many computational models, which have obtained excellent performance on classical classification benchmarks [61]–[64]. Since visual feedback mechanism denotes the procedure of visual perception and cognitive feedback in visual cortex after receiving the visual information stimuli and can enhance the visual spatial and temporal representation ability, Wang et al. [65] integrated bottom-up feature extraction and top-down cognitive feedback in a unified architecture called attentional neural network (ANN) for classification problems, where the top-down influence can deal with high noise or difficult segmentation problems. Cao et al. [66] presented a feedback convolutional neural network (FCNN) where high-level semantic labels can guide the inference of hidden layer neurons' activation status, helping visualize and understand the deep neural networks better, which can be successfully applied to class model visualization, object localization and image classification. Inspired by top-down influences mainly evoked by feedback connections in the human brain [56]–[58], the top-down feedback was used to correct the density prediction of CNN, where a separate top-down CNN accompanies a bottom-up CNN to generate feedback [67]. Shrivastava et al. [14] also applied a top-down feedback signal based on semantic segmentation to Faster region-based CNN (Faster R-CNN), and improved the performance on object detection, semantic segmentation and region proposal generation. Besides, Gatta et al. [68] effectively solved the image parsing problem by operating the top-down semantic feedback in CNN.

Since "what" and "where" pathways in human visual systems have feedback to each other, a computational visual feedback model (VFM) including "what" feedback and "where" feedback [69] was designed between classification and detection for object recognition, where detection windows were rectified by feature saliency for better closeness in the "what" feedback, and object parts were utilized to match object structure for better adaptiveness. In addition to this, [70] designed an attentive feedback network (AFNet) to handle the boundary problem for saliency detection (salient object detection) based on fully convolutional neural networks (FCNs), thus information from encoder module can be employed in corresponding decoder for learning superior segmentations. Chuang et al. [71] come up with a gated-feedback recurrent neural network (GF-RNN) by directing signals from upper recurrent layers to lower ones, which outperforms the traditional methods on character-level language modeling and python program evaluation.

To take full advantage of valuable information among different modalities, a densely cross-level feedback fusion network was put forward for RGB-D salient object detection. This architecture contains two bottom-up streams and a top-down inference pattern, where the former is for learning features, and the latter is for combining cross-modal and cross-level features. Especially, in each level, the cross-modal features will be integrated to generated multiscale RGB-D representations, which are then transmitted to all the shallower layers [72]. Apart from various applications based on feedback mechanism, Lillicrap et al. [62] aimed at reducing computational complexity of the feedback mechanism and proposed a simpler mechanism that replaced 'all the synaptic weights' with 'random synaptic weights, since common backpropagation algorithms in deep learning usually transmit feedback information by multiplying errors with all the synaptic weighs which are not effective enough and considered impossible in the brain. Table 3 summarizes the representative computational models. Although great successes have been achieved by applying feedback strategy to deep neural networks, it is still a challenging problem to take full advantage of neuroscience knowledge for efficient deep learning.

Table 3. Feedback based perception computational models oriented deep learning.

| Methods | Mechanisms | Task description | Learning type |
|---|---|---|---|
| ANN [65] | RBM+ Top-down Feedback | Image Classification | supervised |
| FCNN [66] | CNN+ Top-down Feedback | Object Localization, Image Classification | supervised |
| Top-down CNN[67] | CNN+Top-down Feedback | Crowd Counting | supervised |
| Top-down Faster R-CNN [14] | Faster R-CNN+Top-down Feedback | Object Detection, Semantic Segmentation | supervised |
| TDF-CNN [68] | CNN+Top-down Feedback | Image Parsing | supervised |
| VFM [69] | CNN+'What' Feedback+'Where Feedback' | Robust Object Recognition | supervised |
| AFNet [70] | FCNs+Attentive Feedback | Salient Object Detection | supervised |
| GF-RNN [71] | RNN+Gate-feedback | Character-level Language Modeling | supervised |
| CLF-CNN[72] | CNN+ Cross-level Feedback | RGB-D Salient Object Detection | supervised |

#### 2) Visual Interaction

Visual interaction phenomenon has been proved to exist in visual system and has various manifestations [73]–[75]. Numerous researches about interaction mainly focus on the visual-double stream interaction, and three forms are mostly discussed: "feedback", "continuous cross-talk", and "independent processing" [76]. Besides, [76], [77] point out that different visual information can be interactive and finally finish storage as well as recovery between the dorsal stream and the ventral stream. Since a great amount of evidence certificates that interaction mechanism can enrich visual information and make visual features more distinguishable [78], researchers hope to mimic this mechanism in deep learning for diverse applications.

As one of the crucial interaction forms, the "feedback" mechanism inspired deep learning methods [49], [69] are discussed in detail in the above-mentioned section. Except for the visual feedback-based interaction inspired computational models mentioned before, the "continuous cross-talk" mechanism inspired computational models will also be introduced in this section. As we all know, human can predict the future of a wide range of physical system with just a glance. Inspired by this phenomenon, a visual interaction network (VIN) is proposed to learn the dynamics of physical systems [79]. VIN consists of a perceptual front-end CNN and a dynamics predictor based on interaction networks, which can predict future trajectories of hundreds of time steps with just six input video. Inspired by the multi-stage continuous interaction between the parallel ventral stream and dorsal stream in human brain when recognizing images, Yu et al. design a parallel interaction model (PIM) [80]. Besides, an adaptive visual interaction mechanism is used in DNN for multi-target dynamic state prediction in a real traffic environment [81]. Furthermore, Wei et al. [82], [83] firstly form the continuous interactive completeness of the visual interaction model by incorporating self-interaction, mutual interaction, multi-interaction, and adaptive interaction to it, called visual interaction networks (VIN-Net).

For the "independent processing" interaction form, Krokos et al. apply human visual interaction to variational autoencoder (VAE) and a siamese network to alleviate the burden of intensive manual labeling [84]. Through an interaction iterative process, an analyst can select labeled points and assign them to new or existing communities.

### C. Multi-Modal

In biological system, multimodal interfaces [85] describe interactive systems that seek to leverage natural human capabilities via speech gesture, touch, facial expression, and modalities, bringing more sophisticated pattern recognition and classification methods to human-computer interaction. In human visual system, multimodal [86], [87] refers to the interactive, fusion, and feedback process of different attributes information (text, image, video, etc.) in the visual information transmission procedure.

So far, the intelligent computing model combined with the multi-modal perception of biological vision is a research hotspot in the field of artificial intelligence. Multimodal deep learning has become a major research area with an increasing number of applications that generate/use multiple modalities such as driving, emotion analysis, image fusion, and biometrics. The current multimodal deep learning combined with biological perception can be divided into three types: multimodal representation, multimodal interpretation, and multimodal fusion. Specifically, for multimodal representation, Liu et al. [13] presented a brain-inspired cross-modal neural cognitive computing framework, which is designed for multimedia and multimodal information. In the framework, the hierarchy and other characteristic of architecture and function in the cognitive system have been explored. Mohammad et al. [85] proposed a multimodal perceptual system that is inspired by the inner working of the human brain. The multimodal perceptual system encapsulates parallel distributed processing of real-world stimuli through different sensor modalities and encoding them into features vectors which in turn are processed via a number of dedicated processing units through hierarchical paths. Li et al. [88] preserve the semantic similarity across different modalities and enhance the representation capability of query modality. This framework illustrates the cross-modal retrieval task consists of two main components: deep feature learning and asymmetric hash learning. Yang et al. [89] present a multimodal multi-instance multi-label framework for optimal transport. Inspired by the visual and auditory systems of the human brain, Petar et al. [90] presented a cross-modal convolutional neural network architecture, which can decouple convolutional processing of input partitions.

Thus, for multimodal interpretation, Esma et al. [91] investigated a cross-modal enhancement approach, which is inspired by the auditory information processing in the brain where auditory information is preceded. Su et al. [92] propose a multimodal neural machine transaction framework with semantic interactions between text and image.

Then, for multimodal fusion, motivated by the fact that memory networks consider historical information explicitly, and thus reduce the change of forgetting important historic information, Priyasad et al [93] presented a multimodal data fusion architecture. Cui et al. [94] presented a markov chain recurrent neural network, which can alleviate the item cold start problem by incorporating visual and textural information. Hao et al. [95] argued that the visual-audio modalities should effectively oscillate like those in the brain and the performance the resonance information among visual-audio modalities through capturing the temporal information, and further introducing an internal working memory and external working memory for this audio-visual multi-modality. Table 4 presents the main multimodal perception computational models oriented deep learning.

Table 4. Multimodal perception computational models oriented deep learning.

| Methods | Mechanisms | Task description | Learning type |
|---|---|---|---|
| CNCC [13] | CNN+ text+image+video | Target objection | supervised |
| Multi-DPU [85] | Spiking network +emotion recognition | Emotion recognition | supervised |
| TA-ADCMH [88] | CNN+ text+image | Cross-modal retrieval | unsupervised |
| Multimodal deep network[89] | CNN + text+image | Optimal transport | semi-supervised |
| X-CNN [90] | CNN+ Vision+ audio | Image recognition | supervised |
| Multi-GAT [17] | Graph Network+ multi features | Human activity recognition | transfer learning |
| SER [91] | SNN+ Vision+ audio | Speech emotion recognition | unsupervised |
| Multi modal NMT [92] | RNN+text+image | Multimodal transaction task | supervised |
| MBAF [93] | LSTM+ECG+ SCG | Data classification | supervised |
| MV-RNN [94] | RNN+visual+text | Sequential recommendation | supervised |
| Multimodal model [95] | LSTM +Vision+ audio | Video caption | supervised |

### D. Inference Perception

Visual inference [96] refers to the ability to infer sensory stimuli from predictions that results from internal neural representations built through prior experience. Visual inference perception [97] can capture the potential connections between attributes and goals, and can support us to research high-level structures and mechanisms. Meanwhile, the visual inference can be used to inspire the construction of a variety of computational intelligent models, which can provide a kind of bridge between deep learning and biological visual system. Table 5 presents inference perception computational models oriented deep learning. The current inference perception computational models oriented deep learning is divided into three categories: unsupervised-based, and weakly supervised-based models.

For supervised-based learning with visual inference, Yang et al. [98] proposed a sparse representation based perceptual cognitive and inference process for image quality assessment. Inspired by visual cognition and inference, Vacher et al. [99] present a model with an inference module that is jointly trained with the perception backbone. Specifically, the contrastive learning is used to model the biological visual cognition and reasoning processes, and the paper comes up with two levels of contrast in the model: a novel contrast module and a new contrast loss function. At the model level, the permutation-invariant contrast module is designed to summarize the common features. Inspired by biological inference, Yang et al. [100] presented two pathway-based contour guided visual search framework to salient edge detection. The Bayesian inference is used to auto-weight and integrate the local cues [100]. The biological plausibility of Bayesian inference inspires us to adopt it to combine the contour-based guidance and bottom-up features. The free-energy principle [101] implies that the human visual system (HVS) infers the environment based on the interior states. Inspired by this mechanism, Chen et al. [102] implemented an attention-driven no-reference image quality assessment method with reinforcement learning. Inspired by human biological visual inference scheme, Jiang et al. [103] explored a layer-shippable inference network, which can dynamically carry out coarse-to-fine object categorization. Sancaktar et al. [104] presented a pixel-based deep active inference algorithm, which combines the free energy principle from neuroscience and rooted in variational inference.

For unsupervised-based learning with visual inference mechanism, Xia et al. [105] constructed a deep center-surround inference network and trained it with the data sampled randomly from the entire image to obtain a unified reconstruction pattern for the current image. In this way, an extra inference layer is added at the top to provide ways to explore the center-surround contrast relationship. Inspired by visual hierarchy and inference, Du et al. [106] described a deep generative multi-view framework for neural decoding by employing the fusion of probabilistic modeling and DNNs. The generation and inference procedures in the deep generative model naturally support the cognitive phenomena of imagination. Inspired by human predictive coding theory in human visual cognition system, Xu et al. [107] adopt the encoder-decoder inference architecture to reconstruct the distorted inputs.

Finally, for weakly supervised-based models based on feedforward inference, Cao et al. [108] added the feedforward inference in deep neural network to infer the activation status of hidden layer neurons. In the proposed framework, the feedback network can achieve certain level of selectivity and provide non-relevant suppression during the top-down inference. Moreover, the top-down inference can allow the model to focus on the most salient image regions.

Table 5. Inference perception computational models oriented deep learning.

| Methods | Mechanisms | Task description | Learning type |
|---|---|---|---|
| Sparse represenation [98] | DNN+cognition+inference | image quality assessment | supervised |
| CoPINet [99] | Contrast learning+cognition+inference | Visual inference | supervised |
| CGVS [100] | DNN+ visual search inference | Salient structure detection | supervised |
| NR-IQA [102] | CNN+ free-energy principle | Image quality assessment | supervised |
| Layer-skippable inference [103] | CNN+ visual inference scheme | Object categorization | supervised |
| Deep active inference [104] | CNN + active inference | robot perception and action | supervised |
| C-S inference network [105] | Autoencoder+ inference | Visual saliency estimation | unsupervised |
| DGMM [106] | DNN+ hierarchy+ inference | Reconstruct perceived image | unsupervised |
| BRM-PC [107] | CNN+ visual predict coding | Image quality measurement | unsupervised |
| Look and Think twice [108] | CNN+visual inference | Object localization | weakly supervised |

### E. Specificity and Transfer

In addition to the above-mentioned characteristics of biological perceptual learning, it has two important attributes: specificity and transfer. Specificity can enable the biological system to achieve efficient and robust performance, which includes sparse, visual brightness, contrast, motion, and so on. For instance, in the learning process of biological visual neural network, sparse synaptic connectivity is required for decorrelation and pattern separation in feedforward networks. Transfer is also another important attribute, which means that perceptual learning can be transferred in different positions of the retina. At the same time, it transfers the learning information of the local visual area to another local neural structure [109]. It is obvious that the specificity and transfer also plays an important role in the inspiration of deep neural networks.

For visual specificity, Qiao et al. [110] presented a framework mimicking the visual active and dynamic learning and recognition process of the primate visual cortex. To mimic the visual analogy-detail dual-pathway human visual cognitive mechanism revealed in cognitive science studies, Tao et al [111] proposed a network framework named analogy-detail network for accurate object recognition. The whole network architecture is shown in Fig. 3. The core part of the network is the AD blocks. In the AD cognitive model, the analogy pathway is designed by a lightweight structure:

$$\mathrm{F}_L = f_L\left(L\right) = g\left(v * L\right) \qquad (1)$$

The detail pathway defines the transformation $F_H$ that extract fine features:

$$F_H = f_H\left(H\right) = f_r\left(H\right) + H \qquad (2)$$

where $L$ and $H$ are the corresponding feature maps for low spatial frequency and high spatial frequency signals. $f_L\left(\cdot\right)$, $f_H\left(\cdot\right)$, $v \in R^{C' \times 1 \times 1 \times C}$ is the set of $1 \times 1$ convolutional filters, $f_r\left(\cdot\right)$ are the feature extraction functions. $g\left(\cdot\right)$ is a resize function. Essentially, the analogy pathway can transform $L$ into coarser representation, while the detail pathway can transform $H$ into fine representations. Finally, the coarser and fine representations is fused [111] to produces the outputs $Z$ :

$$M : \left(F_L, F_H\right) \rightarrow Z \qquad (3)$$

To simulated the multi-scale visual processing mechanism of ventral visual stream in the human brain, Zhu et al [112] proposed a multi-scale brain-like network model to predict saliency. Inspired by visual similarity mechanism, Gao et al [113] presented a framework in order to predict perceptual similarity between two texture images. The proposed framework considers both powerful features and perceptual characteristics of contours extracted from the images. The similarity value is computed by aggregating resemblances between the corresponding convolutional layer activations of the two texture maps. To mimic the physiological mechanism of saliency in real human visual system, Wang et al [114] propose ada-sal Network to learn saliency adaptively while feature maps are being trained at the same time. Fernandes et al. [115] design lateral inhibition pyramidal neural network, which is based on the lateral inhibition mechanism of the human visual system.

For visual transfer, Zhang et al. [116] introduced the visual concept transfer mechanism into the deep neural network and constructed a visual concept transfer module. Seitz et al. [117] used a deep neural network for modeling biological visual perceptual learning. They found that the deep neural network reproduced key behavioral results, including increasing specificity with higher task precision. The experimental result also show that the model learns precise discriminations could transfer asymmetrically to coarse discriminations when the stimulus conditions varied. Inspired by the human fundamental transfer learning mechanism, Zhou et al. [118] proposed a visual analogy graph embedded regression model to jointly learn a low-dimensional embedding space and a linear mapping function.

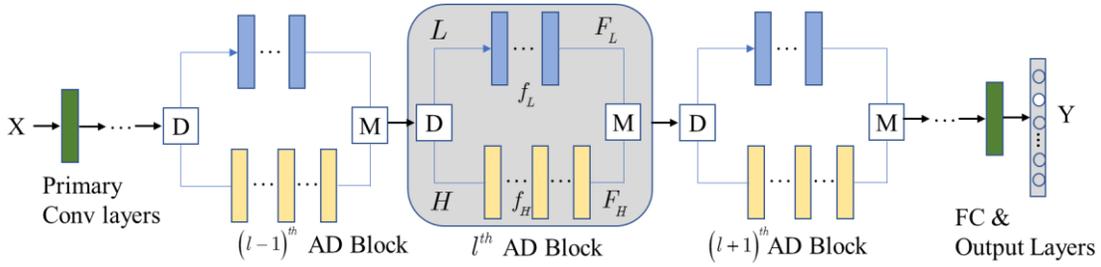

Fig. 3. Visual analogy-detail inspired neural network architecture [111].

## III. THE SERIOUS CHALLENGES FACED BY VISUAL SENSATION AND PERCEPTION COMPUTATIONAL MODEL

With the development of algorithms and technologies such as parallel computing, cloud computing, and artificial intelligence, related technologies of visual perception have been greatly improved whose applications have also taken root in various fields. However, there is a large gap between the machine's visual perception comprehension ability and human visual perception level, it has a wide range of observation, which cannot be observed by the human eye, such as infrared, microwave, and ultrasound. In this section, we analyze the challenges faced by current visual perception computational models oriented deep learning.

### A. Out of Visual Perception Computational Models: Imagination, Planning, Intuitive

Recent perspectives emphasize key ingredients of human intelligence that are already developed in human infants but lacking in most AI systems. Among these capabilities are knowledge of core concepts relating to the physical world, such as space, number, and objectness, which allow people to construct compositional mental models that can guide inferences and prediction. Meanwhile, with the combination of artificial intelligence and biological visual perception system, several most important faculties (intuition, imagination) of the mind need to be considered into the artificial intelligence computational models. Research into human imagination, planning and intuition emphasizes its constructive nature, with human able to construct fictitious mental scenarios by recombining familiar elements in novel ways, necessitating compositional/disentangled representations of the form present in certain generative models [119]. This fits well with the notion that planning in humans involves efficient representations that support generalization and transfer, so that plans forged in one setting can be leveraged in novel environments the share structure. Further, planning and intuition in humans are based on their decisions on logical analysis of all options. For instance, humans seen to plan hierarchically, by considering in parallel terminal solutions, interim, choice points and piecemeal steps toward the goal [120]. These faculties (intuition, imagination, intuition) of the mind are the key issue that researchers must attach great importance during its rapid development. Furthermore, we believe that ultimately these flexible, combinatorial aspects of planning will form a critical underpinning of what is perhaps the hardest challenge for AI research: to build the intelligent computational models that can plan hierarchically, is truly creative, and can generate solutions to challenges that currently elude even the human mind.

### B. Visual Perception Theory

Despite we already know the structure and experimental results of biological visual systems, the visual perception theory is not very well exploited. For instance, we know from many functional magnetic resonance imaging studies in human that there are extensive higher-cognitive functional modulations of the early levels of visual cortex [121]. This naturally leads to the notion that early level representations should be bound to the higher level representations. Meanwhile, these results of functional magnetic resonance imaging can be further developed into visual perception theory.

On the other hand, since biological networks are substantially more complicated than DNNs, it will be even more challenging to understand visual perception mechanisms and structures in the visual systems, then build a relatively complete theoretical system. Much work is still needed to bridge the gap between deep learning and visual perception. In working toward closing this gap, visual perception inspired theory will become indispensable. The relevance of visual perception theory, both as a roadmap for the deep neural network research agenda and as s source of computational tools is particularly salient in the key visual understanding areas.

### C. Architecture Design in Deep Learning

The important deep learning models include recurrent neural network (RNN), auto-encoder (AE), generative adversarial network (GAN), and graph neural network (GNN) are currently widely used in the processing of visual image, industrial time-serial data. Their theoretical problems are mainly reflected in statistics and computing. A simple neural network, such as BP neural network can fit any nonlinear function. Due to the nonlinear liability, deep learning model has better performance for fitting nonlinear function than the simple neural network. However, the represent ability of deep neural network does not represent the model's learnability [122]. That is to say, deep neural network is not intelligent enough, often accompanied by over-fitting and under-fitting problems, and requires the support of big data, but human visual perception system do not complete a large number of calculations to achieve related functions. Meanwhile, visual perception system is distinguished by its ability to rapid learn about new visual information from a handful of examples, leveraging prior knowledge to enable flexible inductive inference [123]. Therefore, to improve and build a more reasonable deep learning network architecture, we need to mimic the mechanisms and theory of visual perception system. Furthermore, we think that deep learning networks can be combined with other biologically inspired intelligent algorithms, such immune algorithms, genetic algorithms, to complete a new level of intelligent computing through hybrid intelligence.

### D. Adaptation, Robustness, and Speed

As visual perception computational models for deep learning have moved forward the state-of-the-art results of various computer vision tasks by a large margin, it becomes more challenging to make progress on top of that. There might be several directions for more powerful models: The first direction is to explore new adaptation of visual perception computational models. Due to the dynamic diversity of actual scenes and the limitation of device computing and storage resources, adaptive computing models based on biological perception are becoming a new design strategy. Meanwhile, deep learning models aim to dynamically adjust either the model structure, the calculation scheme, or both, of them specifically to adapt to the environment context.

A Second direction is to increase the robustness ability of visual perception computational models. Due to the variability of the real world, the quality of visual information is also diverse, and current visual perception and processing technologies in various fields often cannot adapt to such changing visual conditions, such as light intensity and shadows. The low robustness of the algorithm is also a universal problem in this field. Building systems to perform these tasks robustly, with limited computing resources, in real-world scenarios – in the presence of occlusion by objects and other people, changes in illumination and camera pose, variations in the appearance of users, and multiple users – is a huge challenge for the field [122]. A high level of robustness is paramount for practical deployment of these recognition technologies, and in the end robustness can only be determined by thorough testing under a wide range of conditions. To accomplish required tasks at acceptable levels of overall system performance, researchers must determine what the accuracy and robustness requirements are for each component.

The third direction is reducing the computational cost and increase the learning speed. Since the models require a lot of computational resources and are not candidate for real-time application [124], one of the trends toward developing new architecture which allow running the computational models in real-time. The trade-off between speed and accuracy has always been an important issue in the field of visual perception, especially in the field of computer vision. Increasing the processing speed will inevitably reduce the information acquisition and analysis capabilities of deep networks, and vice versa. Its importance is self-evident.

### E. Visual Perception based Integrated Intelligence

In deep learning, integrated learning [125] integrates many neural network structure into a new model. The bio-inspired integrated intelligent computing models have fundamentally difference with the integrated learning in artificial intelligence, which integrates the mechanism, structure, and theory of the biological perception system, and conducts efficient and reasonable model by the simulating the organic integration form of biological visual perception. So far, most of the bio-inspired intelligent computing models only mimic a certain structure or a mechanism of the biological vision system to improve the performance of deep learning models. The disadvantage of these intelligent computing models is that they build the models based on independent biological visual structure or mechanisms. However, the visual perception system is an entirety system that can achieve robust, fast, and accurate visual recognition processes in the complex environments [126]. Therefore, when designing the intelligent computing model, we should consciously the learning type of the biological visual perception system, and consider the architecture design as whole system formation. Furthermore, we can combine the mechanisms, structure of biological vision to build a system-level framework, and finally realize the visual intelligence framework in the biological sense. For instance, visual perception system may use algorithms that exploit specific aspects of neurophysiology- such as dendritic computation, local excitatory-inhibitory network [127], local learning, or other properties- as well as the integrated nature of higher-level visual perception. Such mechanisms or structures promise to allow learning capabilities that go even beyond those of current deep neural networks.

## IV. DEVELOPMENT PROSPECTS OF VISUAL SENSATION AND PERCEPTION COMPUTATIONAL MODELS

Vision is the most important source of information for humans to understand the world. The research pf processing will always accompany human scientific steps. In this section, we introduce its future development directions and trends based on the current challenges of visual perception computational models oriented deep learning, as shown in Fig. 4.

### A. Multi-Modal Multi-Source Information Fusion

Multi-modal fusion by using multiple source of information for visual tasks has exhibited a clear advantage over the unimodal counterpart on various application [128]. Current multi-modal fusion computing models can introduce more biological visual mechanism and theory to further improves the robustness and effectiveness of visual information fusion. For instance, Esma et al. [129] presents ideas and plans for creating a decentralized model for social signal integration inspired by computational models for multi-sensory integration in neuroscience. Besides, choosing the right multi-modal multi-source information fusion techniques for bio-inspired intelligent models is very curial issues. In the further, combining biological visual information, multi-modal multi-source information fusion strategy will still be an important research direction. On the one hand, the multi-modal multi-source information fusion methods can maximize the amount of information. On the other hand, Multi-modal multi-source information fusion methods will be further upgraded to achieve more reliable and accurate results for specific visual perception tasks.

### B. Higher Adaptability and Robustness

Visual perception computational models oriented deep learning will develop towards higher adaptability and robustness of different tasks, which may include transfer learning [130] and meta-learning [131]. For example, the domain adaptation in transfer learning is sub-discipline that deals with the use of models trained on information source distributions in the context of different target distributions. With the learned knowledge transfer to the new task, the performance of the function of domain adaptation is closer to human intelligence. Similarly, meta-learning is an unsupervised learning method, which is intended to allow machines to learn to learn. When the machine has the ability to learn, it can adapt to different visual tasks adaptively and robustly. Meanwhile, these visual perception computational models may be notable for being highly connected and parallel, requiring high-power, and collocating memory and processing. These characteristics provide compelling reasons for developing new hardware that employs visual perception computational models [132].

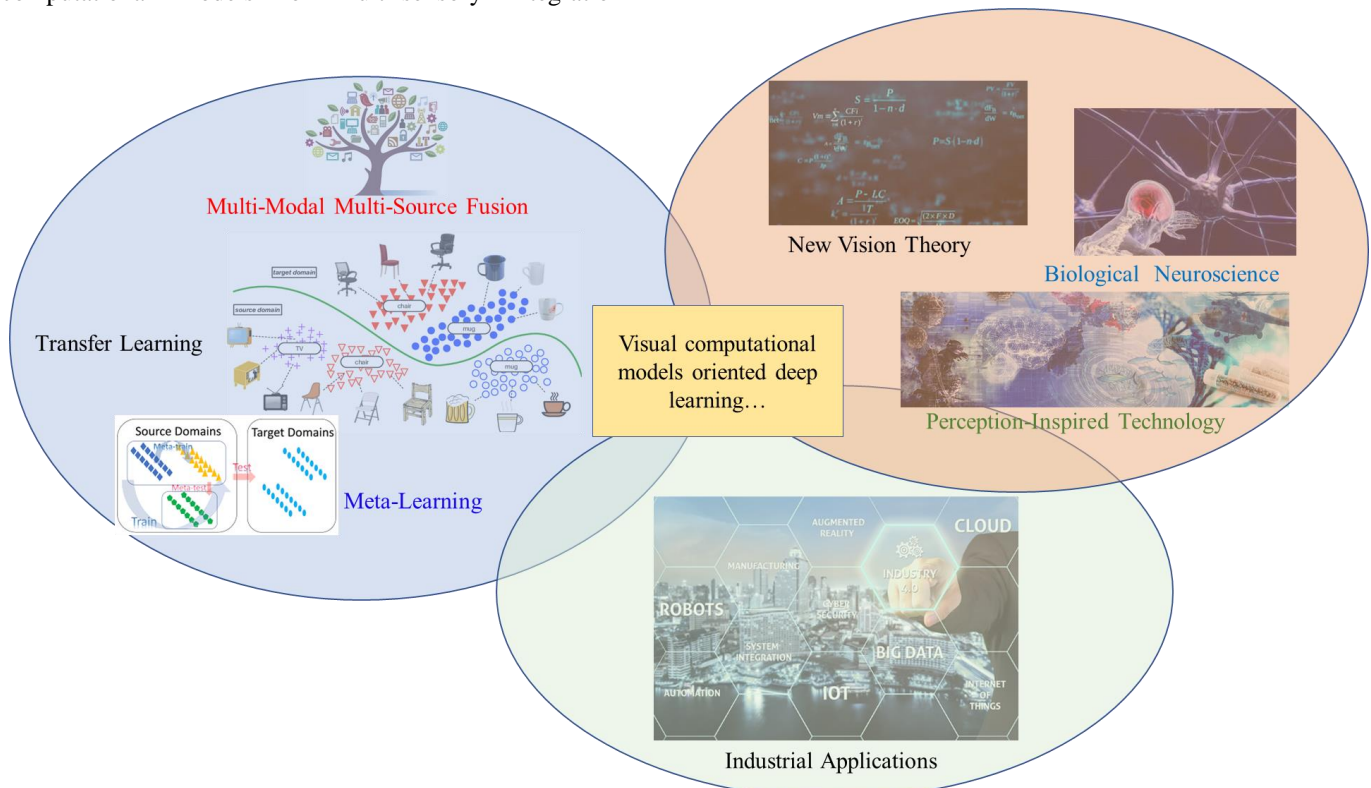

Fig. 4. Development prospects of visual sensation and perception computational models oriented deep learning.

### C.  Industrial Application

In Section II, we have reviewed the visual perception computational models, and detailed the major application filed of computational models, including salient structure detection, Image quality assessment, human activity recognition, industrial time series prediction, industrial soft sensor modeling, and self-driving. These actual tasks and industrial applications basically cover the popular visual perceptions computing research directions in recently years. For example, Qiu et al. [49] development a bio-inspired hierarchical network to understand how the spatiotemporal multivariate time series prediction. The model can automatically reproduce a variety of prediction suppression and familiarity suppression neurophysiological phenomena. Based on the status quo, in the further, most of the bio-inspired visual perception computational models are combined with actual industrial application, which is helpful to human production and life, and has advantages of low cost, high precision and high efficiency.

### D.  From Deep Learning to Neuroscience Perception

With the development of deep learning and neuroscience, the most plausible way forward is to relate the neural networks to the existing well-known physical or biological visual mechanisms and structures. This will aid in developing a meta physical relationship that can help demystify the visual perception system. For instance, Ma et al. [133] explain neural network to the components of a yeast cell, starting with the microscopic nucleotides that make up its DNA, and finally moving to organelles like the mitochondrion and nucleus.

Moreover, the visual perception computational models may also be useful in neuroscience where it can often be difficult to determine, or even quantify, progress. Publicly accessible, large-scale, standardized data-sets are becoming available in neuroscience [134], which may enable the development of neuroscience benchmarks and challenges, for example, to predict the response- properties of neurons along the visual hierarchy, or for comparing representations between artificial networks and the brain. These approaches might be useful in comparing, selecting, and ruling out competing models [135].

### E.  Visual Perception Intelligent Computational Methods to the Trend of World Development.

With the advent of intelligent age, it is an obvious world development trend that the different disciplines intersect and complement each other. Here, visual perception intelligent computational methods will conform to the trend of world development, which will be explained from two main aspects. First, artificial intelligence is a major strategic decision in the layout of various countries in the world, and it is one of the important technologies that affect human development. Also, artificial intelligence need visual perception to guide the engineering of intelligence. Visual perception mechanisms and structures can serve as benchmarks for artificial intelligence, building up form elementary perception abilities to artificial intelligence. Meanwhile, artificial intelligence needs visual perception systems for algorithmic inspiration. Neural network models are an example of a vision-inspired technology that is unrivalled in several domains of artificial intelligence. Second, the research of biological visual perception system is a hot spot in the world, and its development cannot be separated from artificial intelligence to a certain extent. artificial intelligence can be used to expand our understanding of visual perception system, to discover and develop novel therapeutic compounds, to study and characterize visual perception interaction, and to identify design rules and functional network architecture for biological visual synthetic gene circuits [136]. For instance, using existing neural network knowledge to help understand and define the underlying disease mechanisms [137]. Visual perception has always progressed in close interaction with artificial intelligence. The disciplines share the goal of building task-performing models and thus rely on common mathematical theory and programming environments.

## V.  CONCLUSIONS AND FUTURE WORK

Overall, in this paper, we have reviewed and analyzed several major visual perception computational models, including visual attention based, memory based, inference perception based, feedback based, multi-modal based, spatial-time synergy, cross-level and interaction, specificity and transfer, and sparse response. The computational models basically cover the popular visual perception research directions on deep learning in recent years, including CNN, RNN, AE, GAN, Graph neural network, Meta-learning models, and reinforcement learning models. We can conclude that the most of the current visual perception computational models oriented artificial intelligence are helpful to human life and production, and has advantages of low cost, high precision and high efficiency.

In addition, based on the status, we analyze the current challenges faced by humans when using visual perception computational models, including vision acquisition, computing power, device volume, technology security, speed, accuracy, robustness, integrated intelligence, combination, etc. Based on these challenges, we have mode predictions about the development prospects of visual perception. In the future, visual perception will be more closely integrated with artificial intelligence, and will move towards multi-source information fusion, active vision, domain adaptation, meta-learning, reinforcement learning, federal learning, crowd sensing, and other directions, and more field will be applied to visual perception technology.

With the continuous development and intelligentization of visual perception, human production efficiency and quality will continue to improve, which will be one of the important diving forces for human social progress.